%% file: 0_main.tex
\pdfoutput=1

\documentclass[11pt]{article}

\usepackage[final]{acl}
\usepackage{hyperref}
\usepackage{url}
\usepackage{colortbl}
\usepackage{xcolor}
\usepackage{booktabs}
\usepackage{float}
\restylefloat{table}
\usepackage{placeins}
\usepackage{amsmath}

\usepackage{times}
\usepackage{latexsym}

\usepackage[T1]{fontenc}

\usepackage[utf8]{inputenc}

\usepackage{microtype}

\usepackage{inconsolata}

\usepackage{graphicx}

\usepackage{multirow}

\title{Refined Assessment for Translation Evaluation: Rethinking Machine Translation Evaluation in the Era of Human-Level Systems}

\author{
    \textbf{Dmitry Popov\textsuperscript{	$\clubsuit$}},
    \textbf{Vladislav Negodin\textsuperscript{$\clubsuit$}},
    \textbf{Ekaterina Enikeeva\textsuperscript{$\clubsuit$}},
    \\
    \textbf{Iana Matrosova\textsuperscript{$\clubsuit$}},
    \textbf{Nikolay Karpachev\textsuperscript{$\clubsuit$}},
    \textbf{Max Ryabinin\textsuperscript{$\diamondsuit$}}
  \\
  \\
  \textsuperscript{$\clubsuit$} Yandex,
  \textsuperscript{$\diamondsuit$} Together AI
  }

\begin{document}
\maketitle
\begin{abstract}
As machine translation systems approach human-level quality, traditional evaluation methodologies struggle to detect subtle translation errors. We critically examine limitations in current gold-standard approaches (MQM and ESA), including inconsistencies from variable annotator expertise, excessive categorization complexity, coarse severity granularity, accuracy bias over fluency, and time constraints. To address this issue, we introduce a high-quality dataset\footnote{The dataset is available on \href{https://huggingface.co/datasets/yandex/wmt24-en-ru-rate}{Hugging Face}.} consisting of human evaluations for English–Russian translations from WMT24, created by professional linguists. We show that expert assessments without time pressure yield substantially different results from standard evaluations. To enable consistent and rich annotation by these experts, we developed the RATE (Refined Assessment for Translation Evaluation) protocol. RATE provides a streamlined error taxonomy, expanded severity ratings, and multidimensional scoring balancing accuracy and fluency, facilitating deeper analysis of MT outputs. Our analysis, powered by this expert dataset, reveals that state-of-the-art MT systems may have surpassed human translations in accuracy while still lagging in fluency – a critical distinction obscured by existing accuracy-biased metrics. Our findings highlight that advancing MT evaluation requires not only better protocols but crucially, high-quality annotations from skilled linguists.
\end{abstract}

\input{1_intro.tex}
\input{2_RATE}

\input{3_RATE-LLM}
\input{4_related_work}
\input{5_conclusion}

\section{Limitations}

In this section, we discuss the limitations of our research and the proposed RATE protocol.

A comprehensive comparison between RATE, MQM, and ESA protocols would ideally involve the same set of annotators implementing all methodologies under controlled conditions. However, due to the substantial financial resources required for high-skilled annotation, we were unable to conduct such an experiment. While we acknowledge that a direct protocol comparison using the same annotator pool would provide valuable insights, our approach enables meaningful system-level evaluation: the RATE annotations allow derivation of both MQM and ESA quality scores, facilitating comparable system rankings across evaluation frameworks. Importantly, we recognize that the annotation task design itself—including categorization schemes and scoring rubrics—may influence evaluator focus and final scores independent of annotator effects. Given these considerations and cost constraints, we prioritized scaling the dataset size to maximize its utility for both in-depth translation analysis and training future evaluation models.

Our research is confined to English-to-Russian translation evaluation. Although we have no substantial reason to believe the annotation process would differ significantly across other language pairs, the nuanced approach of RATE is particularly valuable for high-resource language directions where machine translation quality closely approximates human translation quality. For languages with more complex orthographic systems or grammatical structures, the error identification process may present additional challenges not encountered in our study.

Regarding reproducibility, human annotation inherently presents challenges as there exists no standardized framework for selecting and qualifying high-skilled annotators. As our findings highlight, the expertise level of human evaluators significantly impacts assessment outcomes. This variability in human resources remains an intrinsic limitation of translation quality evaluation research.

Furthermore, the quality of annotation itself cannot be objectively measured. We must rely on indirect signals such as rigorous annotator selection processes, time spent on annotation, error frequency statistics, and the overlap of identified errors between annotators. Translation evaluation inevitably contains a degree of subjectivity that is extremely difficult to quantify. This inherent subjectivity adds another layer of complexity to establishing definitive benchmarks for translation quality assessment.

\section{Ethics Statement}
ESA and RATE annotators received compensation at standard commercial translation rates for their region, with RATE expert annotators earning double the hourly wage. No personal data was collected during the process, and all presented content was screened for potentially disturbing material. All annotators were informed about how the data would be used prior to beginning the annotation tasks.

Following the annotation, we administered a feedback questionnaire. The vast majority of annotators reported a positive experience and found the instructions clear.

\section*{Acknowledgements}
We extend our deepest gratitude to the expert linguists whose meticulous annotations were indispensable to this research: Tatiana Nazarova, Kseniia Chagina, Iuliia Petrova, Natalia Kharina, Polina Andreeva, Sofia Shishatskaia, Tatiana Slobodeniuk, Alexander Ivolgin, Andrey Poloshak, Irina Skornyakova, Ekaterina Nikiforova, and Anna Andreeva. Their profound language expertise, rigorous attention to detail, and generous investment of time enabled the creation of the high-quality dataset and validation of our RATE protocol. This work would not have been possible without their exceptional contributions to refining translation evaluation standards.

We would also like to sincerely thank Alexandra Godina, Daria Volosova, and the entire Translation Crowd team for their valuable feedback on the annotation instructions and for their contributions to the dataset annotations.

\bibliography{custom}

\appendix

\section{Overlapping ESA Annotation}
\label{sec:appendix_esa_overlap}

To assess the reliability of our annotation scheme, we conducted an additional experiment with overlapping ESA annotations. While the main dataset was annotated using a non-overlapping approach, where each segment was labeled by a single evaluator, we subsequently reannotated the entire dataset twice more. This resulted in each segment having three independent annotations. In this approach, each segment received annotations from three different annotators to mitigate individual preference biases. Each evaluator still annotated all translations of a given source segment.

Analysis of the overlapping annotations revealed that the averaged scores from the three annotators demonstrated reduced variance and thus more stable results compared to single-annotator labeling. This approach provided a more robust estimation of the true quality scores for each segment. Table ~\ref{tab:esa_overlap_3_rankings} presents system rankings for triple-overlapping method. The number of clusters that ESA distinguishes remains unchanged, however the ESA Score ranking is slightly different and a little closer to RATE rankings.

\begin{table}[]
\centering
\renewcommand{\arraystretch}{1.2}  %
\begin{tabular}{lr}
\toprule
ESA Scores ($\uparrow$) & \textbf{ESA} \\
\midrule
\rowcolor{white} Dubformer & 75.91 \\
\rowcolor{white} Claude-3.5 & 74.50 \\
\rowcolor{gray!10} refA & 72.94 \\
\rowcolor{gray!20} Unbabel-Tower70B & 70.27 \\
\rowcolor{gray!40} Yandex & 69.97 \\
\rowcolor{gray!40} GPT-4 & 69.52 \\
\rowcolor{gray!50} ONLINE-G & 67.30 \\
\rowcolor{gray!70} Llama3-70B & 62.80 \\
\end{tabular}

\begin{tabular}{lr}
\toprule
MQM Scores ($\downarrow$) & \textbf{ESA} \\
\midrule
\rowcolor{white} Claude-3.5 & 7.84 \\
\rowcolor{white} Dubformer & 7.99 \\
\rowcolor{gray!10} refA & 10.22 \\
\rowcolor{gray!30} Yandex & 10.23 \\
\rowcolor{gray!30} Unbabel-Tower70B & 10.27 \\
\rowcolor{gray!30} GPT-4 & 10.80 \\
\rowcolor{gray!50} ONLINE-G & 11.61 \\
\rowcolor{gray!70} Llama3-70B & 14.11 \\
\bottomrule
\end{tabular}
\caption{System rankings by ESA and MQM scores. Overlap 3 for ESA.}
\label{tab:esa_overlap_3_rankings}
\end{table}

Additionally, we calculated inter-annotator agreement across our ESA annotation dataset. We computed the agreement by measuring the proportion of simultaneously detected errors, determining the intersection of error spans using the same methodology described in Table \ref{tab:error_overlap}. For each segment, we selected pairs of annotators and calculated the overlap between their annotations. The resulting overlap percentage was 59\%, indicating a reasonably good level of agreement among annotators. This moderate to high agreement suggests that despite the inherent subjectivity in error identification, our annotation guidelines were sufficiently clear to produce consistent results across different annotator groups.

\section{RATE Error Categorization}
\label{sec:rate_error_categorization}
We primarily divide errors into semantic and non-semantic ones. Apart from these terms, we mostly adhere to the MQM terminology for better compatibility, though some of the terms are defined more broadly.

Semantic errors closely correspond to Accuracy in terms of MQM and are split into Mistranslation, Undertranslation, Overtranslation, and Omission. To contrast with them, non-semantic errors encompass issues related to Fluency, Grammar, Style, Inconsistency, Do-not-translate, and Named entities.

In contrast to MQM, our classification does not consider Terminology as an independent type alongside Accuracy, but includes it in Mistranslation. This decision is based on Occam's razor to eliminate redundancy. In addition to this general consideration, there are domain-related reasons as well. It seems justified due to two main reasons. First, wrong terms either distort meaning or not, thereby they are either part of a more general error type (Mistranslation) or relate to another type (False friend, Undertranslation, Fluency, etc.), but there is no reason to identify them as a distinct group. Secondly, it is not evident how the definition of a term can be outlined, whether taken in a strictly scientific meaning or a more casual one. In any case, it is difficult to align annotators on this distinction, which makes the outcome more subjective and noisy.

Undertranslation is combined with Not translated as both categories lead to partial semantic loss, although both match with those in MQM and hardly need more detailed explanation. The same applies to the merging of Overtranslation and Addition.

At the same time, Omission is set as a separate group. At first glance, this may seem contradictory; however, the decision can be easily explained. Both in cases of Undertranslation and Not translated, a reader receives at least part of the original information. It can be reduced (Undertranslation) or a reader can clearly see that a piece of text remains untranslated, but in the case of omission, they have no chance to identify the lack of information.

Certainly, annotators are instructed beforehand that not every omission, under- or overtranslation should be marked as an error, as in many cases appropriate translation requires not maximum but optimal accuracy.

The list of non-semantic errors does not need a profound description. It should only be specified that Grammar also includes Punctuation errors (as syntax is truly a section of grammar) and Fluency is regarded as an extensive category comprising any problems of natural and smooth expression, awkward collocations, tautologies, etc.

The last category here is Other, which is reserved for rare and non-standard cases difficult to categorize.

We deliberately ignore a number of MQM categories which, from our perspective, are irrelevant for the matter of the current study (such as Organization terminology or Third-party terminology) or even lie beyond the borders of translation problems per se (for example, Design and markup or Missing text).

\section{Detailed Statistics on Error Categories}
\label{sec:appendix_error_categories}

Table ~\ref{tab:err_categories_avg_sev} presents the average severity ratings and frequency counts for different translation error categories. The error count represents the average number of errors per text segment. The data reveals a distinct pattern in how different error types impact translation quality. Semantic errors, particularly Mistranslations (3.99), exhibit relatively high average severity, indicating their significant negative impact on translation quality. In contrast, non-semantic errors generally show lower severity values, with style (2.22) and grammar issues (2.38) being perceived as less critical.

Interestingly, while semantic errors are generally considered more severe, certain non-semantic error categories such as inconsistency (4.50) and untranslated content (4.38) received the highest severity ratings overall, despite their relatively low frequency (0.07 and 0.02 per segment, respectively). Fluency errors, though less severe (2.80), were the most frequent individual error category with 1.96 occurrences per segment. Non-semantic errors collectively occurred almost twice as frequently (3.00) as semantic errors (1.56), but their lower average severity (2.74 vs. 3.89) balances the two global error types. The least severe errors fell into the "Other" category (1.93), representing issues that did not fit into the established classification scheme and had minimal impact on overall quality.

\begin{table*}[h]
    \centering
    \begin{tabular}{llcc}
    \toprule
         Error Type & Error Category & Severity & Error Count \\
         \midrule
         \multirow{5}{*}{Semantic} & Mistranslation & 3.99 & 1.16 \\
          & Overtranslation & 3.65 & 0.09 \\
          & Undertranslation & 3.58 & 0.14 \\
          & Omission & 3.58 & 0.17 \\
         \cline{2-4}
          & overall Semantic & 3.89 & 1.56 \\
         \midrule
         \multirow{7}{*}{Non-Semantic} & Inconsistency & 4.50 & 0.07 \\
          & Do Not Translate & 4.38 & 0.02 \\
          & Named Entity & 3.31 & 0.13 \\
          & Fluency & 2.80 & 1.96 \\
          & Grammar & 2.38 & 0.68 \\
          & Style & 2.22 & 0.15 \\
         \cline{2-4}
          & overall Non-Semantic & 2.74 & 3.00 \\ 
         \midrule
         Other & Other & 1.93 & 0.10 \\
    \bottomrule
    \end{tabular}
    \caption{Average number of errors and severity values for different error types and error categories.}
    \label{tab:err_categories_avg_sev}
\end{table*}

\section{Error Types Correlations}
\label{sec:appendix_error_correlations}

The table ~\ref{tab:error_category_correlation} shows Kendall's Tau correlation coefficients between translation error counts per segment and three evaluation metrics (RATE Accuracy, RATE Fluency, and RATE Average, which is average of RATE Accuracy and RATE Fluency). The data is organized by error types (Semantic, Non-Semantic, and Other) and specific error categories within each type. Bold values in the table represent the maximum correlation values in each column, while green highlighting indicates values that are statistically significant from zero at the 0.05 significance level.

The correlation analysis reveals distinct patterns in how different RATE metrics capture specific types of translation errors. The RATE Accuracy demonstrates substantially stronger correlation with semantic errors (-0.65 overall), particularly with Mistranslation (-0.62), highlighting its sensitivity to meaning-related issues in translations. In contrast, the RATE Fluency exhibits markedly higher correlation with non-semantic errors (-0.61 overall), with especially strong correlation to actual Fluency errors (-0.65), confirming its effectiveness in capturing linguistic form and stylistic issues. The RATE Average, which integrates aspects of both accuracy and fluency, shows somewhat stronger correlation with semantic errors (-0.53 overall) compared to non-semantic errors (-0.44), yet maintains robust correlation with both categories. This balanced sensitivity suggests that the RATE Average successfully functions as a comprehensive quality metric, capturing both meaning preservation and linguistic naturalness in translations.

\begin{table*}[h]
    \centering
\begin{tabular}{llrrr}
\toprule
Error Type & Error Category & RATE Average & RATE Accuracy & RATE Fluency \\
\midrule
\multirow{5}{*}{Semantic} & Mistranslation & \cellcolor{green}-0.51 & \cellcolor{green}-0.62 & \cellcolor{green}-0.26 \\
& Overtranslation & \cellcolor{green}-0.10 & \cellcolor{green}-0.15 & -0.03 \\
& Undertranslation & \cellcolor{green}-0.20 & \cellcolor{green}-0.21 & \cellcolor{green}-0.16 \\
& Omission & -0.03 & -0.02 & -0.02 \\
\cline{2-5}
& overall Semantic & \cellcolor{green}\textbf{-0.53} & \cellcolor{green}\textbf{-0.65} & \cellcolor{green}-0.27 \\
\midrule
\multirow{7}{*}{Non-Semantic} & Inconsistency & \cellcolor{green}-0.12 & \cellcolor{green}-0.10 & \cellcolor{green}-0.11 \\
& Do Not Translate & \cellcolor{green}-0.04 & \cellcolor{green}-0.07 & 0.01 \\
& Named Entity & \cellcolor{green}-0.18 & \cellcolor{green}-0.17 & \cellcolor{green}-0.14 \\
& Fluency & \cellcolor{green}-0.43 & \cellcolor{green}-0.18 & \cellcolor{green}\textbf{-0.65} \\
& Grammar & \cellcolor{green}-0.18 & \cellcolor{green}-0.09 & \cellcolor{green}-0.24 \\
& Style & \cellcolor{green}-0.12 & \cellcolor{green}-0.09 & \cellcolor{green}-0.12 \\
\cline{2-5}
& overall Non-Semantic & \cellcolor{green}-0.44 & \cellcolor{green}-0.21 & \cellcolor{green}-0.61 \\
\midrule
Other & Other & \cellcolor{green}-0.10 & \cellcolor{green}-0.09 & \cellcolor{green}-0.11 \\
\bottomrule
\end{tabular}
    \caption{Kendall's Tau correlation between translation error categories and RATE metrics.}
    \label{tab:error_category_correlation}
\end{table*}

\section{Scores Distribution}
\label{sec:scores_distribution}

\begin{figure*}[ht]
    \centering
    \includegraphics[width=0.8\linewidth]{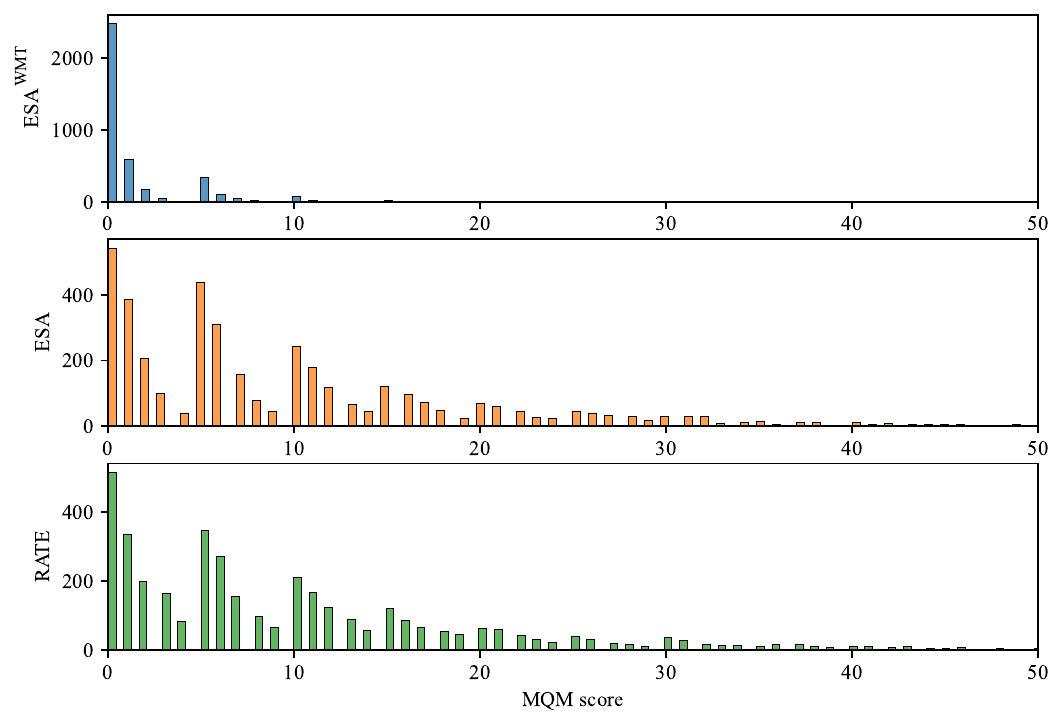}
    \caption{Distribution of MQM scores for ESA\textsuperscript{WMT}, ESA, and RATE.}
    \label{fig:score_distributions_mqm_score}
\end{figure*}

\begin{figure*}[ht]
    \centering
    \includegraphics[width=0.8\linewidth]{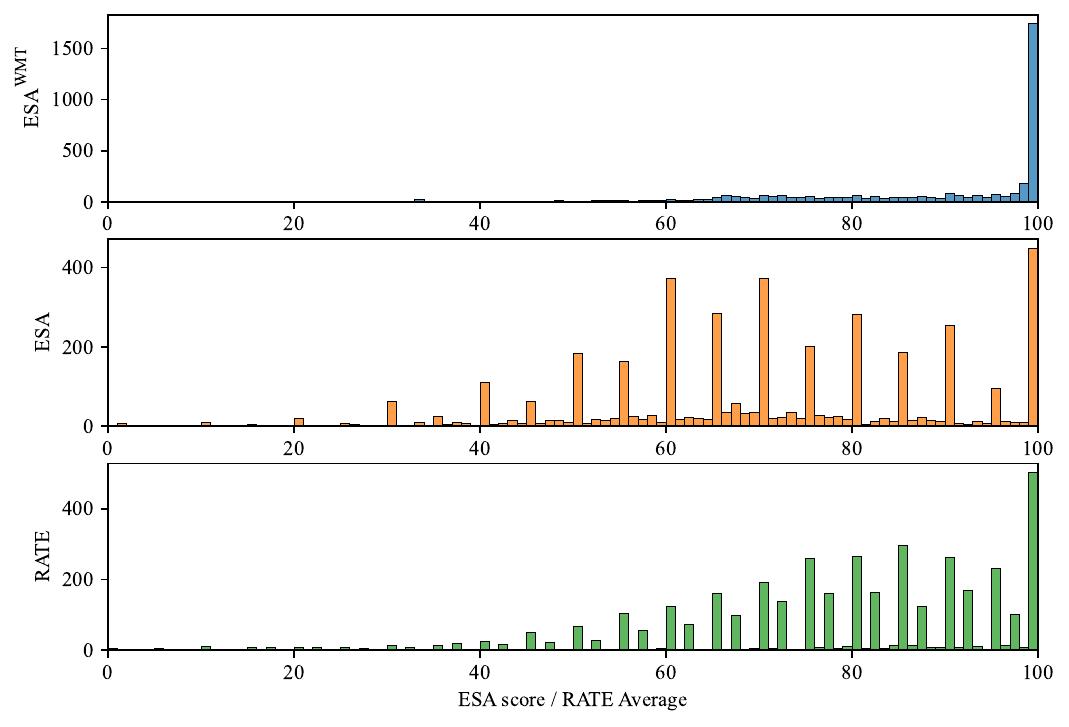}
    \caption{Distribution of ESA scores for ESA\textsuperscript{WMT}, ESA,  distribution of RATE Average.}
    \label{fig:score_distributions_esa_score}
\end{figure*}

Figures ~\ref{fig:score_distributions_mqm_score} and ~\ref{fig:score_distributions_esa_score} illustrate the distribution patterns of translation quality metrics across different evaluation frameworks. In Figure 2, the distribution of MQM scores reveals striking differences between evaluation methods: ESA\textsuperscript{WMT} shows a highly concentrated distribution with most scores clustered near zero, indicating minimal error detection, while both ESA and RATE exhibit more diverse distributions extending up to 50, with gradually decreasing frequencies at higher score values. This pattern suggests that ESA and RATE evaluations demonstrate considerably higher sensitivity to quality variations compared to ESA\textsuperscript{WMT}. The similar shapes of ESA and RATE distributions indicate comparable levels of discriminative power, though with slightly different frequency patterns across the score range.

Figure ~\ref{fig:score_distributions_esa_score} presents the distribution of ESA scores and RATE Average values (RATE Average is an average of RATE Accuracy and RATE Fluency), highlighting further differences in evaluation behaviors. The ESA\textsuperscript{WMT} distribution appears highly polarized with scores concentrated almost exclusively at the top of the scale. In contrast, both ESA and RATE Average distributions show greater utilization of the full scoring range, though with notable peaks at round numbers (particularly visible in the ESA distribution), revealing evaluators' tendency to select psychologically convenient values or to scale reduction. The RATE Average distribution appears somewhat smoother than ESA, because RATE Average represent average of two 0-100 values, which naturally mitigates the "round number effect." These distributions clearly demonstrate how different evaluation methodologies capture translation quality with varying degrees of granularity and sensitivity.

\section{ESA and RATE Framework Interface}
\label{sec:interface}

\begin{figure*}[ht]
    \centering
    \includegraphics[width=1.0\linewidth]{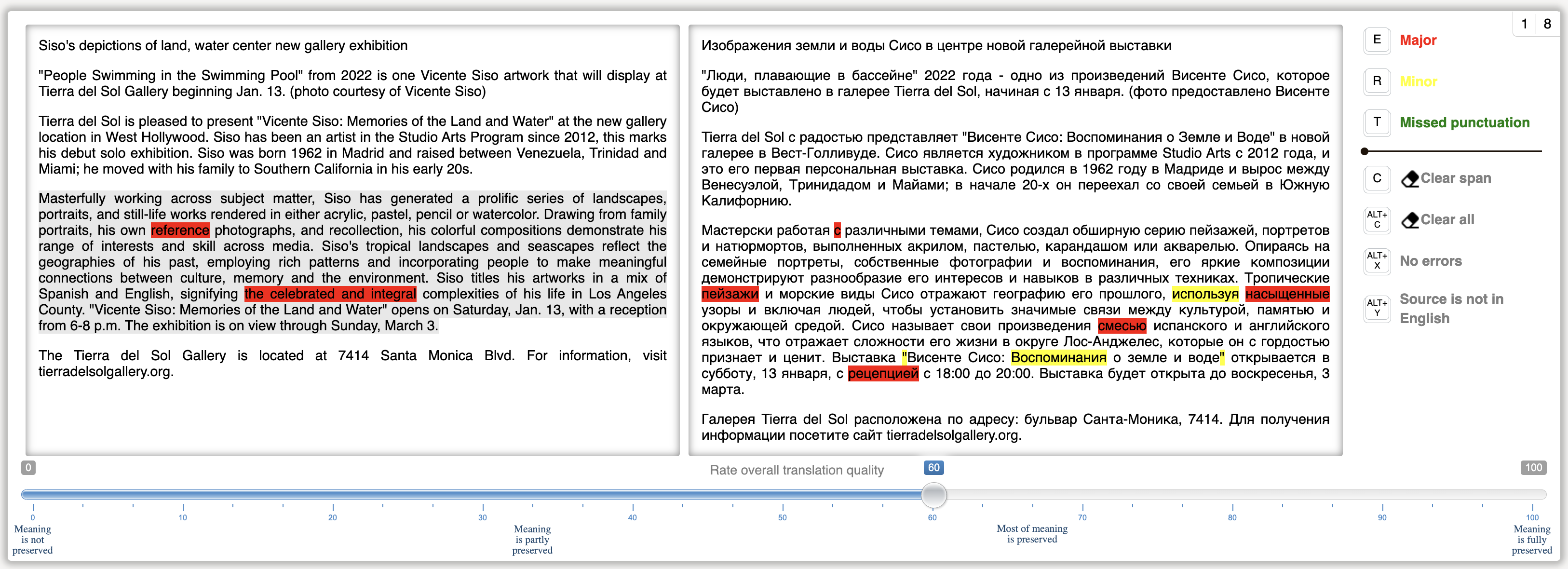}
    \caption{Our ESA interface.}
    \label{fig:esa_interface}
\end{figure*}

\begin{figure*}[ht]
    \centering
    \includegraphics[width=1.0\linewidth]{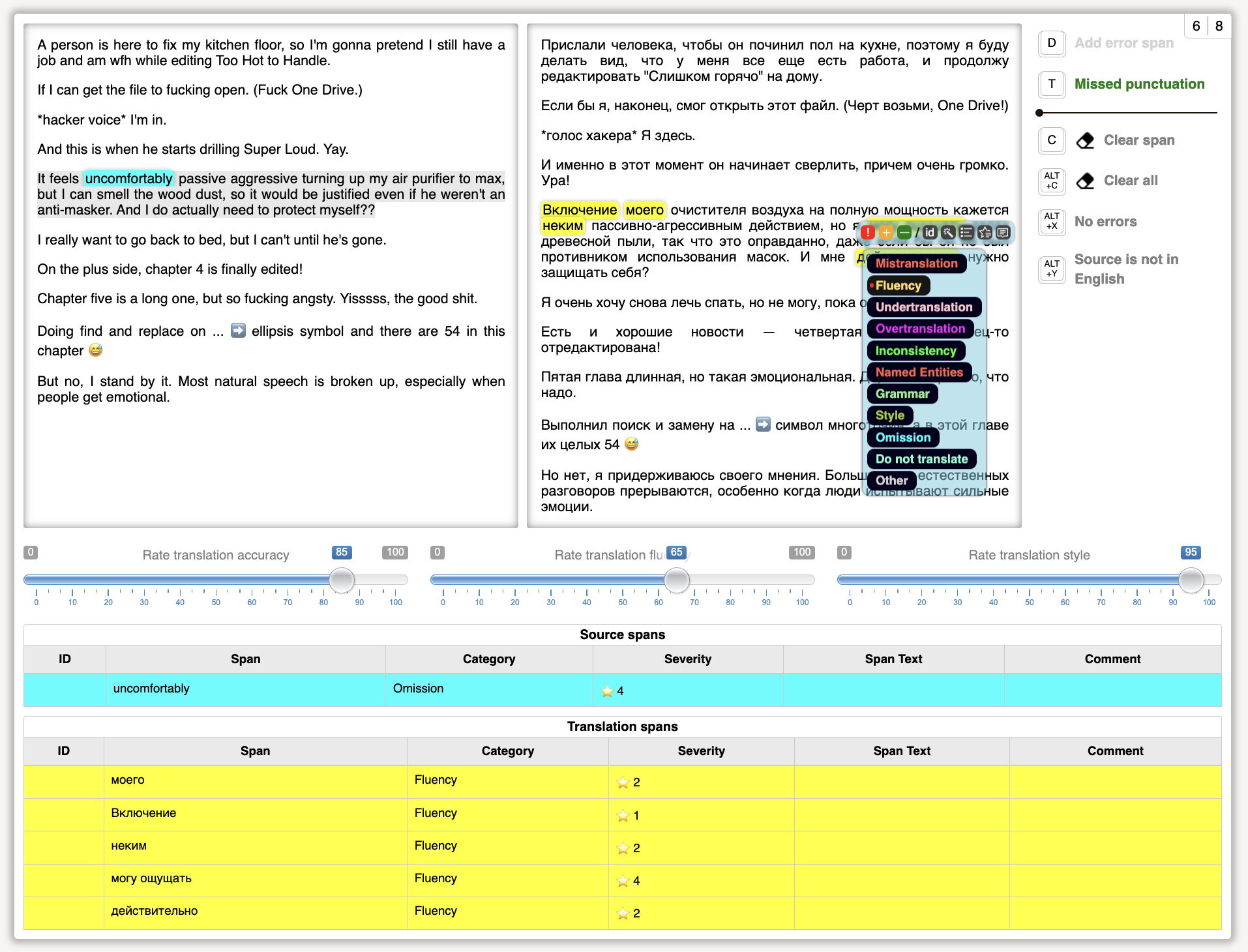}
    \caption{RATE interface.}
    \label{fig:rate_interface}
\end{figure*}

Figures ~\ref{fig:esa_interface} and ~\ref{fig:rate_interface} showcase screenshots of our annotation interfaces designed for ESA and RATE evaluation methodologies, respectively. These visual representations illustrate the distinct approaches to translation quality assessment: the ESA interface provides evaluators with a single 0-100 scale for assigning an overall quality score, while the RATE interface offers a more structured evaluation framework with multiple assessment parameters.

The ESA interface allows evaluators to mark only two types of errors—major and minor—while the RATE interface provides significantly more granularity through dropdown menus where annotators can specify error categories, severity levels, add comments, and link spans across the text by ID to clarify their annotations (e.g. in case of inconsistent translation). This enhanced categorization in RATE enables more detailed qualitative analysis beyond simple error counts.

Both interfaces permit spans to be highlighted in both the source text and the translation. The ability to highlight portions of the source text serves a critical function, allowing annotators to indicate content that has been omitted in the translation—representing a more sophisticated alternative to the [MISSING] token from the original ESA methodology. This feature provides a clear visual indication of content omissions that might otherwise be difficult to annotate in the target text alone.

For both systems, annotators are instructed to focus exclusively on the segment highlighted in gray and its corresponding translation, ensuring consistent evaluation scope across different methodologies. Our implementation also introduces a specialized annotation for missing punctuation, which is automatically classified as a minor error in ESA, while in RATE it is assigned to the Grammar category. This standardization allows for more systematic comparison between the evaluation frameworks while maintaining their distinct characteristics.

\section{RATE Annotation Instruction}
\label{sec:rate_instruct}

\begin{figure*}[ht]
    \centering
    \includegraphics[width=1.0\linewidth]{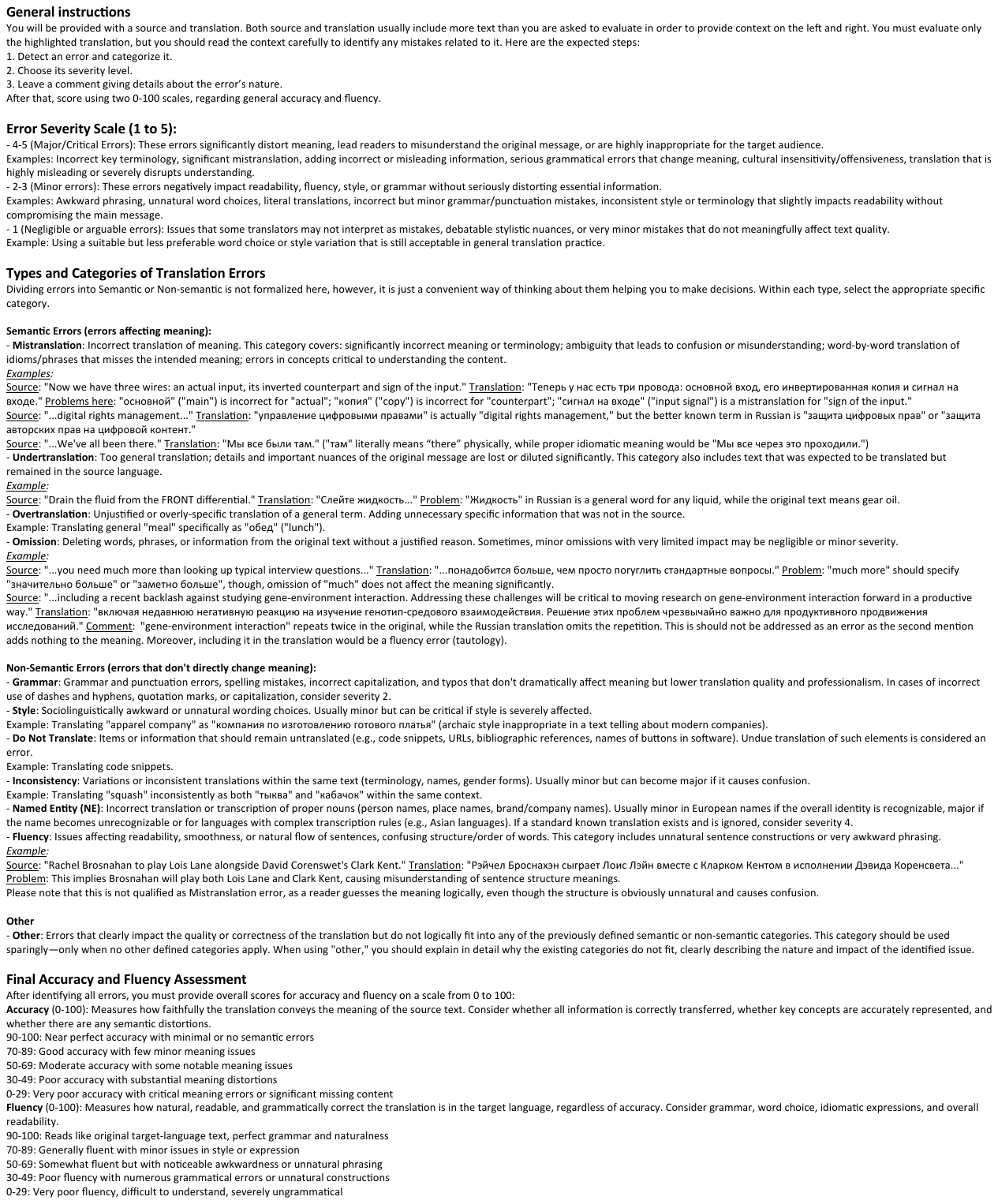}
    \caption{RATE annotation instruction.}
    \label{fig:rate_annotation_instruction}
\end{figure*}

Figure ~\ref{fig:rate_annotation_instruction} shows RATE annotation instruction.

\end{document}

%% file: 1_intro.tex
\section{Introduction}

Recent advances in natural language generation have made evaluation increasingly challenging, as modern systems often produce outputs that match or exceed human-written references \citep{clark2021, gemini}. 
For generative AI applications, human feedback has become the cornerstone of progress, with techniques like Reinforcement Learning from Human Feedback driving significant improvements \citep{ouyang2022}. 
However, the process of collecting human judgments faces numerous challenges: annotators seek shortcuts to simplify evaluation tasks \citep{ipeirotis2010}, often prioritizing surface-level properties over aspects requiring deeper analysis. 
Despite these known issues, the field continues to rely heavily on human evaluation, particularly for open-ended generation tasks, where annotators typically provide single overall scores or preference rankings with limited transparency into their decision-making processes \citep{novikova2018}.

Machine translation (MT) represents a particularly well-established field where human evaluation plays a critical role. 
Over decades, the community has developed sophisticated frameworks for translation quality assessment, with the Multidimensional Quality Metrics (MQM) \citep{mqm, freitag2021experts} framework and its simplified derivative Error Span Annotation (ESA) \cite{esa} emerging as dominant paradigms. 
These evaluation standards are reinforced through annual Conference on Machine Translation (\href{https://www2.statmt.org/wmt24/index.html}{WMT}) campaigns, which produce benchmark datasets that influence the majority of contemporary research on translation quality assessment. 
Indeed, the field's progress is largely measured through improvements against these benchmarks.

Despite this established infrastructure, translation quality assessment remains fundamentally challenging. It demands high levels of cognitive effort, linguistic expertise in multiple languages, domain knowledge, and substantial time investment, all of which conflicts with the practical constraints of large-scale evaluation campaigns.
Our analysis reveals concerning limitations in current gold standard translation evaluation protocols, including excessive complexity of error categorization in comprehensive approaches like MQM, an oversimplified binary (major/minor) error severity classification, significant bias toward accuracy at the expense of fluency, and alarmingly brief annotation times (often under one minute per segment).

The contributions of this paper include: (1) a critical analysis of current MT evaluation practices, with empirical evidence of their limitations; (2) new high-quality annotations for the WMT24 dataset that demonstrate the impact of improved evaluation methodologies on system rankings and enable fine-grained analysis of differences between systems that cannot be distinguished using original WMT'24 annotations; (3) the RATE framework for more comprehensive translation quality assessment that streamlines error categorization while providing balanced measurement of both accuracy and fluency dimensions. Through these contributions, we aim to stimulate discussion about evaluation standards that have remained largely unchallenged despite the rapid evolution of translation systems.

%% file: 2_RATE.tex
\section{Current Gold Standard Approaches and Their Limitations}

\subsection{MQM and ESA}\label{sec:mqm_esa}

The Multidimensional Quality Metrics (MQM) \cite{mqm, freitag2021experts} and Error Span Annotation (ESA) \cite{esa} have emerged as dominant paradigms for human evaluation of machine translation quality. 
MQM provides a hierarchical framework of error categories, with errors classified into major types (Accuracy, Fluency, Style, Terminology, etc.) and further sub-categorized into more specific error types. 
Each error is typically weighted according to its severity (minor, major, or critical), and these weights contribute to an overall quality score. 
This approach allows for detailed error analysis but requires significant training and time investment from annotators.

ESA, a simplified version of MQM, focuses on identifying and categorizing spans of text containing errors, classifying them primarily as major or minor, with less emphasis on detailed error typology. While annotators mark error spans, the comparison between translation systems actually relies on the Direct Assessment (DA) score obtained during annotation rather than directly utilizing the identified errors. This DA score, which we will refer to as ESA score throughout this paper, ranges from 0 to 100, with interface guidelines that help annotators assess meaning preservation (0\% - No meaning preserved, 33\% - Some meaning preserved, 66\% - Most meaning preserved, 100\% - Perfect). Major errors are defined as those that seriously change meaning, make text difficult to read, or decrease usability. This approach was designed to reduce annotation complexity while still providing meaningful quality assessment, combining continuous rating with high-level error severity span marking.

Both frameworks have been widely adopted in translation evaluation campaigns such as the Conference on Machine Translation (WMT), serving as gold standards for machine translation systems comparison and development.

\subsection{Limitations of the Existing Approaches}
\paragraph{Problematic Error Categorization in MQM} The MQM framework suffers from several interconnected issues that compromise its effectiveness. 
Its core version (8 categories, 36 subcategories) is already unwieldy, with the full version expanding to approximately 200 subcategories—creating excessive cognitive burden for annotators. 
Despite this expansiveness, it paradoxically lacks coverage for common issues (e.g., tautologies in languages like Russian and French) while simultaneously containing redundant categories with overlapping definitions (e.g., "Date format" vs. "Date/time", or "Ambiguous target/source content" and "Unclear reference"). 
Additionally, many error types are placed in counterintuitive categories (e.g., "Unjustified euphemism" under Mistranslation rather than Style). 
These structural problems force annotators to make arbitrary categorization decisions, potentially reducing inter-annotator agreement and compromising evaluation reliability.

\paragraph{Single-score Evaluation Inadequacy} Condensing translation quality into a single numerical score proves increasingly problematic for differentiating modern MT systems \citep{flamich2025feedbirdsscoreaccuracynaturalness}. 
Translation quality encompasses multiple distinct dimensions (accuracy, fluency, stylistic appropriateness), and different domains prioritize these aspects differently. 
A single aggregate score inevitably introduces bias toward certain dimensions while obscuring others, making it difficult to meaningfully distinguish between systems or provide targeted improvement guidance. 
As MT quality approaches human parity, more nuanced multi-dimensional evaluation becomes essential.
\paragraph{Major/Minor Distinction Lacks Granularity} The MQM score construction relies heavily on weighting counts of major and minor errors. 
This binary severity classification proves excessively reductive in practice. 
Translation errors exist on a continuum of severity: some errors significantly exceed the threshold of ``minor'' without reaching the impact level of ``major.'' 
This coarse granularity fails to capture the subtle quality differences between advanced translation systems and potentially misrepresents the true user experience with translated content.
\paragraph{Opaque Human Evaluation Processes} Documentation of human evaluation methodology in major benchmarks often lacks transparency regarding crucial details. 
While evaluators are frequently described simply as ``experts,'' specific information about their qualifications, training protocols, or assessment guidelines remains scarce. 
Even more concerning is the limited time allocated for assessment. 
Recent studies \citep{esa, wmt24} report median annotation times of just 34 seconds for ESA and 49 seconds for MQM evaluations per segment per system --- durations that appear insufficient for a thorough analysis of nuanced translation issues. 
Such constraints potentially compromise assessment quality, particularly when evaluating high-performing systems where errors may be subtle and require careful attention to detect.

The central challenge we face is that high-quality translation evaluation has evolved into a complex task even for human judges. Meaningful quality assessment that can effectively distinguish between advanced translation systems and provide actionable insights requires both highly qualified evaluators with sufficient time and a refined methodology that addresses the limitations of current approaches.

\section{RATE: Refined Assessment for Translation Evaluation}

This section of our study presents an attempt to design a more effective evaluation system that provides informative quality reports while adhering to five core principles: comprehensiveness, consistency, comparability with established metrics, compactness, and convenience for both human annotators and automated analysis.

The proposed RATE workflow consists of three main stages: error categorization, severity determination, and quality scaling.

\paragraph{Error Categorization} Our RATE protocol preserves error categorization as a critical component for providing actionable translation feedback while significantly streamlining the taxonomy. 
Rather than eliminating categories entirely (as in ESA) or maintaining the extensive hierarchy of MQM, we adopt a balanced approach that distinguishes between semantic errors (affecting meaning: Mistranslation, Undertranslation, Overtranslation, Omission) and non-semantic errors (affecting form: Grammar, Fluency, Style, Inconsistency, Named entities, Do-not-translate).

This simplified classification enables annotators to provide specific, diagnostic information about translation weaknesses without the cognitive burden and redundancy of MQM's extensive category system. 
For instance, we incorporate terminology errors within Mistranslation, as they fundamentally represent meaning distortions, and we reserve an "Other" category for exceptional cases. Detailed category definitions appear in Appendix~\ref{sec:rate_error_categorization}.

\paragraph{Severity Assessment} RATE expands the traditional binary (major/minor) severity classification to a 5-point scale. 
The scale ranges from 5 (critical errors with severe impact on meaning or readability) down to 1 (minimal issues that even native speakers might overlook). 
This expanded scale provides the necessary granularity to differentiate between slight inaccuracies and consequential mistranslations. 
The middle point (3) captures moderate errors that fall between the traditional major and minor categories.

\paragraph{Quality Scaling} The final assessment stage involves two distinct quality dimensions, each rated on a 1-100 scale. 
Accuracy measures faithfulness to source meaning, while fluency assesses grammatical correctness and natural expression. 
This multidimensional approach allows evaluators to characterize systems with different strengths and weaknesses, such as translations achieving high accuracy but using stilted language, or reading naturally while introducing subtle meaning shifts.

\paragraph{RATE Score and RATE MQM Score} RATE Score quantifies translation quality by summing the severity values of all identified errors, providing a comprehensive measure of error impact. 
For compatibility with established metrics, we also calculate RATE MQM Score by converting our expanded severity ratings into the traditional binary classification, treating higher severity errors (values 4 and 5) as major and moderate errors (values 2 and 3) as minor, then applying the standard formula: $5 \times \text{(number of major errors)} + \text{(number of minor errors)}$.

\paragraph{Implementation Considerations} The RATE protocol strikes a balance between the excessive complexity of MQM and the potential oversimplification of ESA approaches. 
It provides sufficient detail for actionable insights while maintaining annotator efficiency. 
However, effective implementation requires more than just protocol documentation. 
The framework relies on high-skilled annotators with strong bilingual proficiency, and requires proper training on error categorization, severity calibration, and quality scaling. 
Appendix~\ref{sec:rate_instruct} provides detailed calibration guidelines for severity levels and quality scales, along with concrete examples for each error category to ensure consistent application of the framework.

\section{Data Collection and Annotation}

\subsection{Selection of Translation Direction} Our study focuses exclusively on English-to-Russian translation. 
English-Russian represents a high-resource direction where modern systems have achieved impressive performance, making it ideal for examining the more subtle challenges in evaluating near-human-quality translations.

While WMT evaluates numerous language directions, resource constraints inevitably limit the depth of analysis for any single pair. 
By concentrating on this specific direction, we enable the thorough linguistic assessments that high-quality translation evaluation demands.

\subsection{Annotator Selection and Qualification} Following our concerns about the expertise required for high-quality translation assessment, we implemented a rigorous annotator selection process, recruiting professional translators with linguistics or translation degrees, substantial industry experience, and verified language proficiency.
Candidates demonstrated their skills through translation post-editing tests that required context analysis and fact-checking, followed by interviews.
This multi-stage process yielded 13 highly qualified specialists who received comprehensive training on the RATE framework prior to evaluation work.

For comparison purposes, we also collected ESA-style annotations from a larger group of in-house evaluators.
While these annotators were native Russian speakers with verified C1 English proficiency, we did not require specific academic degrees or translation experience.
This group received standard training for ESA and other conventional annotation schemes.

Annotators received average professional translator's or evaluator's payment.

\subsection{Dataset and Experimental Setup}
\begin{table}
\centering
\begin{tabular}{lrrr}
\toprule
Domain &  \#Docs &  \#Segments &  \#Tokens \\
\midrule
literary &     7 &        63 &    2494 \\
news &    16 &        94 &    4984 \\
social &    33 &       274 &    4847 \\
speech &    66 &        66 &    4954 \\
\bottomrule
\end{tabular}
\caption{Annotated dataset statistics}
\label{tab:data_stats}
\end{table}

We utilized the WMT'24 dataset described in \citet{wmt24}, which contains texts from four domains: news (17 documents), literary (8 documents), social (34 documents), and speech (111 documents).
The original dataset includes 13 system translations for the English-Russian language pair.
Since our focus is on evaluating high-quality translations, we selected 8 systems for our experiment: the human reference (refA) and the 7 top-rated systems.

To balance domain representation, we subsampled segments (corresponding to document paragraphs) to include up to 9 segments per document from news, literary, and social domains, and 1 segment per document from the speech domain, which originally contained single-segment texts. This sampling strategy yielded a total of 3976 annotated segment-system pairs for expert evaluation. The resulting dataset structure is presented in Table~\ref{tab:data_stats}. Our dataset is publicly available on \href{https://huggingface.co/datasets/yandex/wmt24-en-ru-rate}{Hugging Face}.

\subsection{Annotation Setups}
The RATE annotation setup follows the same segment-level paradigm used in MQM and ESA evaluations, where each annotator examines all translations of a given source segment.

\paragraph{ESA-style Annotation} Performed by our control group of annotators. 
We extended the annotation guidelines with examples of major and minor errors specific to English-to-Russian translation. 
Each annotation task presented the source text with the highlighted segment and 8 translations, with the entire document available for context. 
In other respects, the annotation process and interface conformed to the WMT'24 setup.
Throughout our analysis, we refer to this annotation as \textbf{ESA}, while the standard WMT annotation is denoted as \textbf{ESA\textsuperscript{WMT}}.

\paragraph{RATE Annotation by Expert Translators} Our primary expert group performed annotations using the RATE framework. 
As in the previous setup, annotators evaluated 8 translations simultaneously. 
Following our protocol, they highlighted and categorized error spans and provided overall accuracy and fluency scores for each segment.

\section{Results}
\subsection{Annotator Qualification Impact on Evaluation}

Our findings demonstrate the substantial impact of annotator qualifications and expertise on translation evaluation outcomes.

\paragraph{Annotation Time} A striking disparity exists between annotation groups as evidenced in Table~\ref{tab:annotation_times} RATE annotators spent a median time of 6.31 minutes per segment, approximately 11 times longer than ESA\textsuperscript{WMT} annotators (34 seconds). 
Even our ESA annotators, working in a nearly identical setup to ESA\textsuperscript{WMT} but with stronger qualification requirements, spent significantly more time (2.46 minutes) than WMT annotators.

\paragraph{Error Detection} This time investment directly correlates with error identification rates shown in Table~\ref{tab:avg_errors}. 
RATE annotators identified approximately 7 times more errors per segment than ESA\textsuperscript{WMT} annotators (4.66 vs 0.65), while our ESA annotators detected 5 times more errors (3.34 vs 0.65). 
Notably, the 7× difference between our RATE and ESA\textsuperscript{WMT} annotations mirrors findings from previous studies \citep{esa} comparing high-quality MQM\textsuperscript{WMT} annotations with standard ESA.

\paragraph{Annotator Quality vs. Protocol} Dramatic differences between our ESA implementation and ESA\textsuperscript{WMT}, despite identical protocols, underscore that annotator selection may be even more critical than the evaluation methodology itself.
\begin{table}
\centering
\begin{tabular}{lccc}
\toprule
 & ESA\textsuperscript{WMT} & ESA & RATE \\
\midrule
Literary & - & 2.88 & 8.93 \\
News & - & 3.57 & 14.39 \\
Social & - & 1.80 & 3.84 \\
Speech & - & 3.53 & 12.76 \\
\midrule
Overall & 0.57 & 2.46 & 6.31 \\
\bottomrule
\end{tabular}
\caption{Median annotation times per segment per system (minutes)}
\label{tab:annotation_times}
\end{table}

\begin{table}
\centering
\begin{tabular}{lccc}
\toprule
 & \#Errors & \#Majors & \#Minors \\
\midrule
ESA\textsuperscript{WMT} & 0.65 & 0.22 & 0.43 \\
ESA & 3.34 & 1.68 & 1.66 \\
RATE & 4.66 & 1.74 & 2.59 \\
\end{tabular}
\begin{tabular}{lccccc}
\toprule
Severity & 5 & 4 & 3 & 2 & 1 \\
\midrule
RATE & 0.75 & 0.99 & 1.25 & 1.33 & 0.33 \\
\bottomrule
\end{tabular}
\caption{Average number of errors per segment across different evaluation methods}
\label{tab:avg_errors}
\end{table}

\begin{table*}[h]
\centering
\renewcommand{\arraystretch}{1.2}  %
\begin{tabular}{lc}
\toprule
ESA Scores ($\uparrow$) & \textbf{ESA\textsuperscript{WMT}} \\
\midrule
\rowcolor{white} Dubformer & 90.49 \\
\rowcolor{white} Unbabel-Tower70B & 89.41 \\
\rowcolor{white} refA & 89.05 \\
\rowcolor{white} Claude-3.5 & 88.49 \\
\rowcolor{white} Yandex & 86.80 \\
\rowcolor{white} GPT-4 & 85.67 \\
\rowcolor{white} ONLINE-G & 85.59 \\
\rowcolor{gray!70} Llama3-70B & 78.23 \\

\end{tabular}
\begin{tabular}{lc}
\toprule
ESA Scores ($\uparrow$)& \textbf{ESA} \\
\midrule
\rowcolor{white} Dubformer & 76.01 \\
\rowcolor{white} Claude-3.5 & 75.53 \\
\rowcolor{gray!10} refA & 73.00 \\
\rowcolor{gray!20} Yandex & 70.60 \\
\rowcolor{gray!40} Unbabel-Tower70B & 70.56 \\
\rowcolor{gray!40} GPT-4 & 69.99 \\
\rowcolor{gray!50} ONLINE-G & 68.00 \\
\rowcolor{gray!70} Llama3-70B & 63.24 \\

\end{tabular}
\begin{tabular}{lc}
\toprule
ESA Scores ($\uparrow$) & \textbf{RATE} \\
\midrule
\rowcolor{white} refA & 83.94 \\
\rowcolor{white} Dubformer & 83.87 \\
\rowcolor{gray!10} Claude-3.5 & 80.60 \\
\rowcolor{gray!20} Unbabel-Tower70B & 78.56 \\
\rowcolor{gray!30} Yandex & 78.34 \\
\rowcolor{gray!40} GPT-4 & 76.47 \\
\rowcolor{gray!50} ONLINE-G & 74.37 \\
\rowcolor{gray!70} Llama3-70B & 70.17 \\

\end{tabular}

\begin{tabular}{lr}
\toprule
MQM Scores ($\downarrow$) & \textbf{ESA\textsuperscript{WMT}} \\
\midrule
\rowcolor{white} refA & 0.99 \\
\rowcolor{white} Claude-3.5 & 1.06 \\
\rowcolor{white} Unbabel-Tower70B & 1.14 \\
\rowcolor{white} Dubformer & 1.23 \\
\rowcolor{gray!10} GPT-4 & 1.45 \\
\rowcolor{gray!40} Yandex & 1.47 \\
\rowcolor{gray!40} ONLINE-G & 1.96 \\
\rowcolor{gray!70} Llama3-70B & 3.06 \\
\bottomrule
\end{tabular}
\begin{tabular}{lc}
\toprule
MQM Scores ($\downarrow$) & \textbf{ESA} \\
\midrule
\rowcolor{white} Claude-3.5 & 7.56 \\
\rowcolor{white} Dubformer & 7.60 \\
\rowcolor{gray!10} Yandex & 9.93 \\
\rowcolor{gray!30} refA & 10.01 \\
\rowcolor{gray!30} Unbabel-Tower70B & 10.09 \\
\rowcolor{gray!30} GPT-4 & 10.34 \\
\rowcolor{gray!50} ONLINE-G & 11.23 \\
\rowcolor{gray!70} Llama3-70B & 13.73 \\
\bottomrule
\end{tabular}
\begin{tabular}{lc}
\toprule
MQM Scores ($\downarrow$) & \textbf{RATE} \\
\midrule
\rowcolor{white} Dubformer & 8.21 \\
\rowcolor{gray!10} Claude-3.5 & 9.27 \\
\rowcolor{gray!20} refA & 9.38 \\
\rowcolor{gray!30} Unbabel-Tower70B & 10.99 \\
\rowcolor{gray!40} Yandex & 11.15 \\
\rowcolor{gray!50} GPT-4 & 12.24 \\
\rowcolor{gray!60} ONLINE-G & 13.11	 \\
\rowcolor{gray!70} Llama3-70B & 16.02 \\
\bottomrule
\end{tabular}
\caption{System rankings by ESA and MQM scores. Systems with the same background color are not statistically significantly different.}
\label{tab:esa_mqm_rankings}
\end{table*}

\subsection{System Ranking Comparison}
\begin{table*}
\centering
\renewcommand{\arraystretch}{1.2}  %
\begin{tabular}{lr}
\toprule
RATE Accuracy ($\uparrow$) & \\
\midrule
\rowcolor{white} Dubformer & 83.65 \\
\rowcolor{white} Claude-3.5 & 82.39 \\
\rowcolor{gray!20} refA & 80.81 \\
\rowcolor{gray!30} GPT-4 & 78.12 \\
\rowcolor{gray!40} Unbabel-Tower70B & 76.06 \\
\rowcolor{gray!50} Yandex & 73.87 \\
\rowcolor{gray!70} ONLINE-G & 71.48 \\
\rowcolor{gray!70} Llama3-70B & 71.36 \\
\bottomrule
\end{tabular}
\begin{tabular}{lr}
\toprule
RATE Fluency ($\uparrow$) & \\
\midrule
\rowcolor{white} refA & 87.08 \\
\rowcolor{gray!10} Dubformer & 84.08 \\
\rowcolor{gray!20} Yandex & 82.82 \\
\rowcolor{gray!30} Unbabel-Tower70B & 81.05 \\
\rowcolor{gray!40} Claude-3.5 & 78.80 \\
\rowcolor{gray!50} ONLINE-G & 77.26 \\
\rowcolor{gray!60} GPT-4 & 74.81 \\
\rowcolor{gray!70} Llama3-70B & 68.98 \\
\bottomrule
\end{tabular}
\begin{tabular}{lr}
\toprule
 RATE Scores ($\downarrow$)&  \\
\midrule
\rowcolor{white} Dubformer & 11.25 \\
\rowcolor{white} refA & 12.15 \\
\rowcolor{white} Claude-3.5 & 12.56 \\
\rowcolor{gray!10} Yandex & 13.97 \\
\rowcolor{gray!25} Unbabel-Tower70B & 14.19 \\
\rowcolor{gray!40} GPT-4 & 15.87 \\
\rowcolor{gray!55} ONLINE-G & 16.16 \\
\rowcolor{gray!70} Llama3-70B & 19.55 \\
\bottomrule
\end{tabular}

\caption{System rankings by RATE Accuracy, RATE Fluency scores and RATE scores.}
\label{tab:accuracy_rankings}
\end{table*}

A common outcome of human evaluation in machine translation is an aggregated quality score per system, enabling direct comparison. As described in Section~\ref{sec:mqm_esa}, the MQM score is derived as a weighted sum of major and minor error counts per segment-system pair, while the ESA score (DA score) represents the direct assessment of translation quality.
Crucially, the rich annotations collected via the RATE protocol allow us to compute both the MQM score and the ESA score for each segment-system pair, ensuring compatibility with these established benchmarks.
For the RATE-derived ESA score, we calculated it by averaging the overall Accuracy and Fluency scores assigned during RATE annotation.

Table~\ref{tab:esa_mqm_rankings} presents system rankings according to different evaluation protocols. Systems within the same gray-shaded cluster do not differ significantly from each other based on Wilcoxon rank-sum tests.

\paragraph{ESA Score Rankings} The ESA\textsuperscript{WMT} annotation corresponding to our subsample identifies only two statistically distinct clusters, with one cluster containing 7 of the 8 systems (Table~\ref{tab:esa_mqm_rankings}). 
This limited differentiation likely stems partly from our dataset's smaller size compared to the complete WMT'24 dataset, though it's worth noting that even on the full dataset, ESA\textsuperscript{WMT} adds just one additional cluster that only separates the human reference (refA). 
Nevertheless, the improved discriminative power of our protocols is evident: RATE distinguishes 7 clusters while our ESA implementation identifies 6 clusters among the same systems.

\paragraph{MQM Score Rankings} The contrast becomes even more pronounced when examining MQM-based rankings (Table~\ref{tab:esa_mqm_rankings}). 
Here, ESA\textsuperscript{WMT} identifies 4 clusters, our ESA implementation distinguishes 5 clusters, and RATE achieves complete separation with 8 distinct clusters (one for each system). 
This increasing granularity in system differentiation highlights how qualified annotators with sufficient time allocation can make more nuanced quality distinctions, even when working with similar evaluation frameworks.

\paragraph{Multidimensional Evaluation Insights} Comparing translation systems based on varied evaluation protocols and metrics produces inconsistent rankings, making it difficult to draw definitive conclusions about which system performs best. 
This inconsistency highlights a fundamental challenge: compressing multiple translation qualities into a single score inevitably introduces bias toward specific aspects, resulting in contradictory interpretations depending on the evaluation framework used.

Both ESA and MQM frameworks demonstrate this issue through their strong accuracy orientation. 
The ESA protocol explicitly instructs annotators to evaluate "meaning preservation" (essentially accuracy), while research on MQM \cite{freitag2021experts} has shown that scores are "primarily driven by major and accuracy errors, as most major errors involve accuracy issues" (a finding also confirmed in our study, Table~\ref{tab:acc_fl_avg_sev}).

This accuracy bias is historically justified and understandable: preserving meaning has traditionally been considered the primary and most fundamental requirement of translation. 
However, this focus creates an incomplete picture of translation quality. 
Our results in Table~\ref{tab:accuracy_rankings} show that machine translation systems, which have only recently made significant breakthroughs in this dimension, now surpass human reference in accuracy metrics, a remarkable achievement that would have seemed impossible just a few years ago. 
Yet it's increasingly evident that meaning preservation, while essential, represents just one aspect of high-quality translation.

Interestingly, while machine translation systems have surpassed the human reference in accuracy dimensions, the human translation maintains clear superiority in fluency evaluations. 
This distinction reveals an important nuance in the evolving narrative around MT quality: claims that "machine translation has surpassed human translation" \cite{wmt24} oversimplify a complex reality where different systems excel in different dimensions of translation quality. 
This multidimensional perspective also suggests that in certain domains, such as literary works, social media content, or other informal communications, optimal translation might actually benefit from prioritizing fluency over perfect accuracy, a trade-off that current evaluation metrics struggle to accommodate.

\begin{figure*}[ht]
    \centering
    \includegraphics[width=1.0\linewidth]{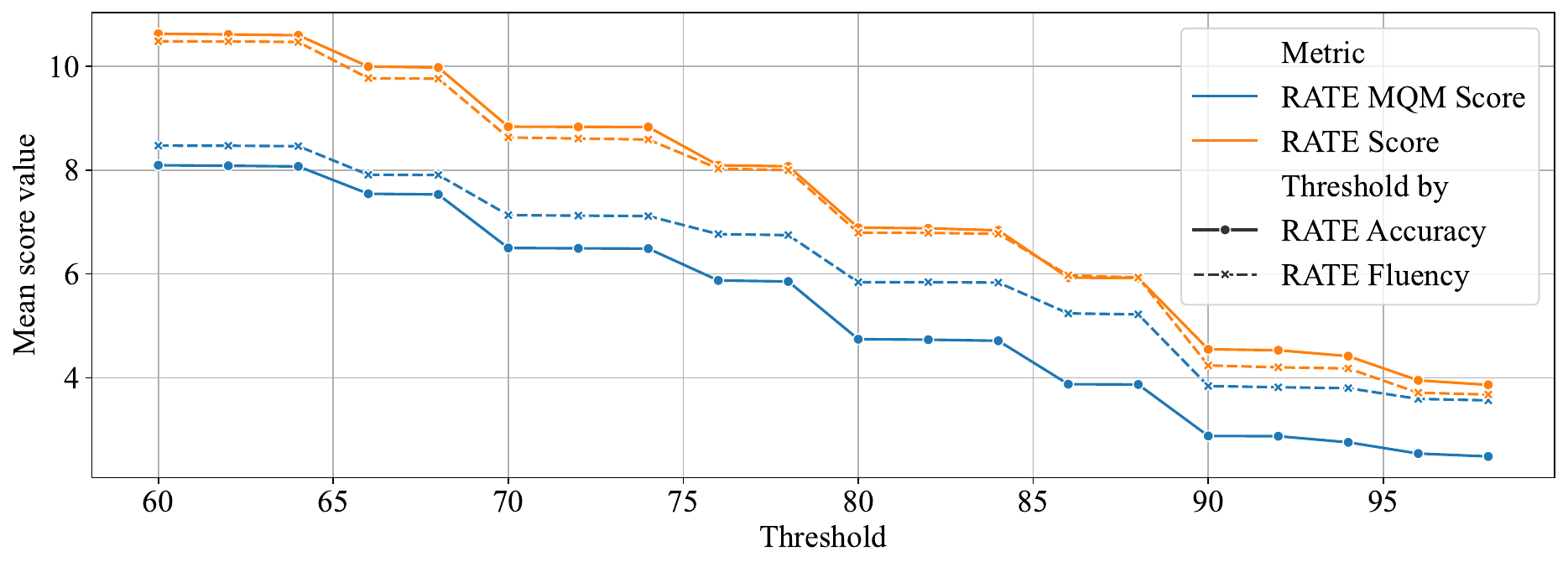}
    \caption{Dependence of mean MQM and Severity scores on the fluency/accuracy threshold. For each threshold, only examples with values above the threshold are taken into account.}
    \label{fig:mqm_rate_threshold}
\end{figure*}

The RATE Score we propose offers a more balanced approach by integrating various translation aspects more effectively than existing metrics. 
Figure~\ref{fig:mqm_rate_threshold} provides empirical evidence of this balance, showing how MQM and RATE Scores respond to different quality thresholds. 
The graph plots the mean score values when considering only examples above specific accuracy or fluency thresholds. 
The MQM curve (blue) consistently tracks closer to the accuracy threshold line (solid) than to the fluency threshold line (dashed), demonstrating its stronger sensitivity to accuracy aspects. 
The gap between MQM's response to accuracy versus fluency thresholds is particularly noticeable in the higher range (80--100). 
In contrast, the RATE Score (orange) maintains more comparable distances from both the accuracy and fluency threshold lines throughout the entire range, confirming its more balanced sensitivity to both dimensions. 
This balanced measurement approach places all three top-performing systems (refA, Dubformer, and Claude-3.5) in a single statistical cluster, indicating comparable overall quality despite their dimensional differences when evaluated holistically.

This approach enables a two-stage evaluation strategy: first, identify systems with strong overall translation quality using the RATE Score, and then analyze their specific dimensional differences. 
Such an analysis reveals that while state-of-the-art MT systems excel in accuracy, human translation maintains a significant advantage in fluency, information that would be obscured in a single-dimensional evaluation framework.

\begin{table}
\centering
\begin{tabular}{lcc}
\toprule
 & Severity & Error count \\
\midrule
Semantic & 3.89 & 1.56 \\
Non-semantic & 2.74 & 3.00 \\
\bottomrule
\end{tabular}
\caption{Average number of errors and severity values for different error types}
\label{tab:acc_fl_avg_sev}
\end{table}

\subsection{Error Span Overlap Analysis}

\begin{table}
\centering
\renewcommand{\arraystretch}{1.2}  %
\begin{tabular}{lccc}
\toprule
\rule{0pt}{2.8ex}$\downarrow AB \rightarrow$ & ESA\textsuperscript{WMT} & ESA & RATE \\
\midrule
\rule{0pt}{2.5ex}ESA\textsuperscript{WMT} & - & 25\% & 21\% \\
ESA & 88\% & - & 52\% \\
RATE & 90\% & 65\% & - \\
\midrule
XCOMET-XXL & 81\% & 36\% & 33\% \\
GEMBA-MQM & 85\% & 49\% & 44\% \\
\bottomrule
\end{tabular}
\begin{tabular}{l}
$|A \cap B|/|B|$ \quad (100\% for $B = \emptyset$)
\end{tabular}
\caption{Overlap of error spans between different evaluation methods. Table shows proportion of errors identified by method B were also found by method A}
\label{tab:error_overlap}
\end{table}
To understand the relationship between different annotation approaches, we analyzed the overlap between error spans identified by each method.

Table~\ref{tab:error_overlap} presents overlap percentages between annotation approaches, revealing significant differences in error detection patterns. 
RATE annotations capture 90\% of errors identified by ESA\textsuperscript{WMT}, while ESA\textsuperscript{WMT} covers only 21\% of RATE-identified errors—suggesting RATE annotators found the same errors as WMT assessors plus many additional issues. 
Similarly, RATE identified 65\% of ESA errors, while ESA captured only 52\% of RATE errors, showing that highly skilled professional translators in the RATE process detected more subtle translation issues than those using the ESA methodology.

We also evaluated two automatic error detection systems: XCOMET-XXL \cite{xcomet} and GPT-4o based GEMBA-MQM \cite{gemba_mqm}. 
While these systems demonstrated better span overlap than ESA\textsuperscript{WMT} (81-85\% coverage of ESA\textsuperscript{WMT} errors), they still underperformed compared to human ESA annotators in detecting errors identified by other methods (covering only 36--49\% of ESA errors and 33--44\% of RATE errors). These results highlight the continuing gap between automated and human evaluation systems, though automated approaches show promising improvements over some human evaluation protocols.

%% file: 3_RATE-LLM.tex
\relax

%% file: 4_related_work.tex
\section{Related Work}

Over the past decades, human evaluation of machine translation (MT) has evolved significantly. Early assessments relied on holistic scoring \cite{ALPAC, ARPA}, gradually giving way to more structured approaches. The Conference on Machine Translation (\href{https://www2.statmt.org/wmt24/index.html}{WMT}), established in 2006, has become the primary venue for standardizing evaluation methodologies \cite{koehn-monz-2006-manual}, shifting from ranking-based assessments to Direct Assessment (DA) \cite{graham-etal-2013-continuous} , where annotators assign continuous quality scores. The Multidimensional Quality Metrics (MQM) framework \cite{mqm, freitag2021experts} introduced a comprehensive error typology that classifies issues by type and severity. ESA, a simplified version of MQM, focuses on identifying and categorizing spans of text containing errors, classifying them primarily as major or minor, with less emphasis on detailed error typology \cite{esa}.

Though WMT has been instrumental in collecting human judgments at scale, such evaluations remain expensive and time-consuming. Error-marking protocols like MQM, despite being labor-intensive, provide actionable insights crucial for system improvement that holistic scoring cannot offer. These practical constraints have motivated automated metrics like COMET \cite{comet, xcomet}, BLEURT \cite{bleurt}, and MetricX \cite{metricx}, with recent work exploring Large Language Models for evaluation \cite{gemba_mqm, lu-etal-2024-error, zhang-etal-2025-good, lu-etal-2025-mqm}. Error spans identified through LLM annotation \cite{gemba_mqm} have been incorporated into AI-assisted human evaluation frameworks, effectively reducing costs while maintaining evaluation quality. However, all these automated approaches ultimately depend on human-curated gold standards for validation. Currently, high-quality MQM annotations exist at scale primarily for English–German and Chinese–English pairs only, creating a bottleneck for progress as top MT systems converge in quality, making fine-grained distinctions increasingly important yet difficult to assess.

%% file: 5_conclusion.tex
\section{Conclusion}
Our study introduces the RATE framework for translation evaluation and contributes a high-quality annotation dataset for English-to-Russian translations. We demonstrate that existing gold standard methods show significant bias toward accuracy at the expense of fluency, while RATE offers a more balanced single-score metric that weighs both dimensions, complemented by granular statistics for comprehensive system analysis.

Our findings reveal that annotator qualifications and time investment dramatically impact evaluation outcomes, potentially more than the protocol itself. We also discovered that state-of-the-art MT systems have surpassed human translations in accuracy while still lagging in fluency, a nuance obscured by one-dimensional evaluation approaches. The RATE balanced methodology is an important advancement in translation quality assessment, particularly as MT systems continue to improve and require more sophisticated evaluation methods.